\documentclass[10pt,twocolumn,letterpaper]{article}

\usepackage{cvpr}              % To produce the CAMERA-READY version
 \usepackage{tabularx}%

\usepackage[numbers]{natbib}

\usepackage{graphicx}
\usepackage{amsmath}
\usepackage{amssymb}
\usepackage{booktabs}
\usepackage{hhline}
\usepackage{subcaption}
\usepackage{dsfont}
\usepackage{scrextend}
\usepackage[dvipsnames]{xcolor}
\usepackage{pifont}
\usepackage{enumitem}

\usepackage[pagebackref,breaklinks,colorlinks]{hyperref}

\newcommand{\ua}{$\uparrow$}
\newcommand{\da}{$\downarrow$}
\newcommand{\bb}[1]{\textbf{#1}}

\newcommand{\cmark}{\ding{51}}%
\newcommand{\xmark}{\ding{55}}%

\usepackage[nameinlink,capitalize]{cleveref}
\crefname{section}{Sec.}{Secs.}
\Crefname{section}{Section}{Sections}
\Crefname{table}{Table}{Tables}
\crefname{table}{Table}{Tabs.}
\crefname{figure}{Fig.}{figures}

\def\cvprPaperID{8361} % *** Enter the CVPR Paper ID here
\def\confName{CVPR}
\def\confYear{2022}

\begin{document}

%%%%%%%%% TITLE - PLEASE UPDATE
\title{More than Words: In-the-Wild Visually-Driven Prosody for Text-to-Speech}

% \author{First Author\\
% Institution1\\
% Institution1 address\\
% {\tt\small firstauthor@i1.org}
% % For a paper whose authors are all at the same institution,
% % omit the following lines up until the closing ``}''.
% % Additional authors and addresses can be added with ``\and'',
% % just like the second author.
% % To save space, use either the email address or home page, not both
% \and
% Second Author\\
% Institution2\\
% First line of institution2 address\\
% {\tt\small secondauthor@i2.org}
% }

\author{
Michael Hassid\\
Google Research\\
{\tt\small hassid@google.com}\and
Michelle Tadmor Ramanovich\\
Google Research\\
{\tt\small tadmor@google.com}\and
Brendan Shillingford\\
DeepMind\\
{\tt\small shillingford@deepmind.com}\and
Miaosen Wang\\
DeepMind\\
{\tt\small miaosen@deepmind.com}\and
Ye Jia\\
Google Research\\
{\tt\small jiaye@google.com}\and
Tal Remez\\
Google Research\\
{\tt\small talremez@google.com}
}

\maketitle

%%%%%%%%% ABSTRACT
\begin{abstract}

%%%%%%%%%%%%%%%%%%%%%%%%%%%% INTRO STUFF BELOW %%%%%%%%%%%%%%%%%%%%%%%%%%%%%%%%%
In this paper we present VDTTS, a \textbf{V}isually-\textbf{D}riven \textbf{T}ext-\textbf{t}o-\textbf{S}peech model. Motivated by dubbing, 
VDTTS takes advantage of video frames as an additional input alongside text, and generates speech that matches the video signal.
%
%%%%%%%%%%%%%%%%%%%%%%%%%%% DETAILS/EVAL %%%%%%%%%%%%%%%%%%%%%%%%%%%%%%%%
%
% We demonstrate how this allows VDTTS to, unlike plain TTS models, generate speech with prosodic variations, like natural pauses and pitch.
We demonstrate how this allows VDTTS to, unlike plain TTS models, generate speech that not only has prosodic variations like natural pauses and pitch, but is also synchronized to the input video.

Experimentally, we show our model produces well-synchronized outputs, approaching the video-speech synchronization quality of the ground-truth, on several challenging benchmarks including ``in-the-wild'' content from VoxCeleb2.
%
%
% Furthermore, we demonstrate that the text and speaker embedding supply the speech content, while the prosody is produced by the video signal.
% %
% Our novel system is trained and evaluated on a range of difficult open-domain filming conditions, with a wider vocabulary than available in other common datasets.
% %
%Through rigorous evaluation, w
% We demonstrate that our method has superior performance compared to other works in terms of automated metrics and human listening studies.
%
%We achieve near ground-truth quality on the GRID dataset, and on open-domain YouTube data from VoxCeleb2, a 2.45 mean opinion score compared to 2.79 of the ground-truth when raters were asked to evaluate video-speech synchronization.
% According to raters asked to evaluate video-speech synchronization, we achieve near ground-truth quality on both the GRID dataset and on more difficult open-domain YouTube data from VoxCeleb2, with a 2.45 mean opinion score compared to 2.79 of the ground-truth.
%\tr{what does the MOS represent?}
% old:
% We encourage the reader to view the supplementary demo videos at the project page 
% \footnote{ \url{google-research.github.io/lingvo-lab/vdtts}}, demonstrating video-speech synchronization, robustness to speaker ID swapping, and prosody.
% new:
Supplementary demo videos demonstrating video-speech synchronization, robustness to speaker ID swapping, and prosody, presented at the project page.\footnote{\label{project_page}Project page: \newline \url{http://google-research.github.io/lingvo-lab/vdtts}}

%{\small \url{http://google-research.github.io/lingvo-lab/vdtts}}

%\bs{what's up with the URL font?}
%  \url{https://PROJECT_FUTURE_URL},
% In this paper we present VDTTS, a visual-driven TTS model. VDTTS takes advantage of an additional silent video as an input to generate speech with pauses, emotions, prosody and pitch, that match the video signal. Our method is composed of video and text encoders that are combined via a multi-source attention layer. Speech is generated by a mel-spectrogram decoder followed by a vocoder. We evaluate our method on several challenging benchmarks including VoxCeleb2. Unlike recent automatic voice over systems, this is the first time such a method is trained and evaluated on in-the-wild examples that include unseen speakers. 
% %
% Through rigorous evaluation we demonstrate the superior performance of our method with respect to other recent work both in terms of objective measures as well as human listening studies. \todo{maybe we can put some numbers here.}

\end{abstract}

% In this paper we present VDTTS, a \textbf{V}isually-\textbf{D}riven \textbf{T}ext-\textbf{t}o-\textbf{S}peech model. Motivated by dubbing, 
%VDTTS takes advantage of video frames as an additional input alongside text, and generates speech with prosody that matches the video signal.
% The model produces near ground-truth quality on the GRID dataset.
% On open-domain ``in-the-wild'' evaluations, the model produces well-synchronized outputs approaching the video-speech synchronization quality of the ground-truth.
% We show VDTTS performs favorably compared to alternate approaches.

% % KEY HERE: put bad/shortcomings near good
% Intriguingly, VDTTS is able to produce video-synchronized speech without any explicit losses or constraints to encourage this, suggesting complexities such as synchronization losses or explicit modeling are unnecessary.
% Furthermore, we demonstrate that the text and speaker embedding supply the speech content, while the prosody is produced by the video signal.
% Our results also suggest that the ``easiest" solution for the model to learn is to infer prosody visually, rather than modeling it from the text.

%%%%%%%%% BODY TEXT
\section{Introduction}
\label{sec:intro}
Post-sync, or dubbing (in the film industry), is the process of re-recording dialogue by the original actor in a controlled environment after the filming process to improve audio quality. Sometimes, a replacement actor is used instead of the original actor when a different voice is desired such as Darth Vader's character in Star Wars~\citep{nyt}.

Work in the area of automatic audio-visual dubbing often approaches the problem of generating content with synchronized video and speech by (1) applying a text-to-speech (TTS) system to produce audio from text, then (2) modifying the frames so that the face matches the audio~\citep{yang_large-scale_2020}.
The second part of this approach is particularly difficult, as it requires generation of photorealistic video across arbitrary filming conditions.

In contrast, we extend the TTS setting to input not only text, but also facial video frames, producing speech that matches the facial movements of the input video.
The result is audio that is not only synchronized to the video but also retains the original prosody, including pauses and pitch changes that can be inferred from the video signal, providing a key piece in producing high-quality dubbed videos. 

% Besides dubbing-like applications, our method naturally extends to other applications such as low-quality speech enhancement in videos, audio restoration in captioned videos.

\begin{figure}[t]
\includegraphics[width=0.47\textwidth]{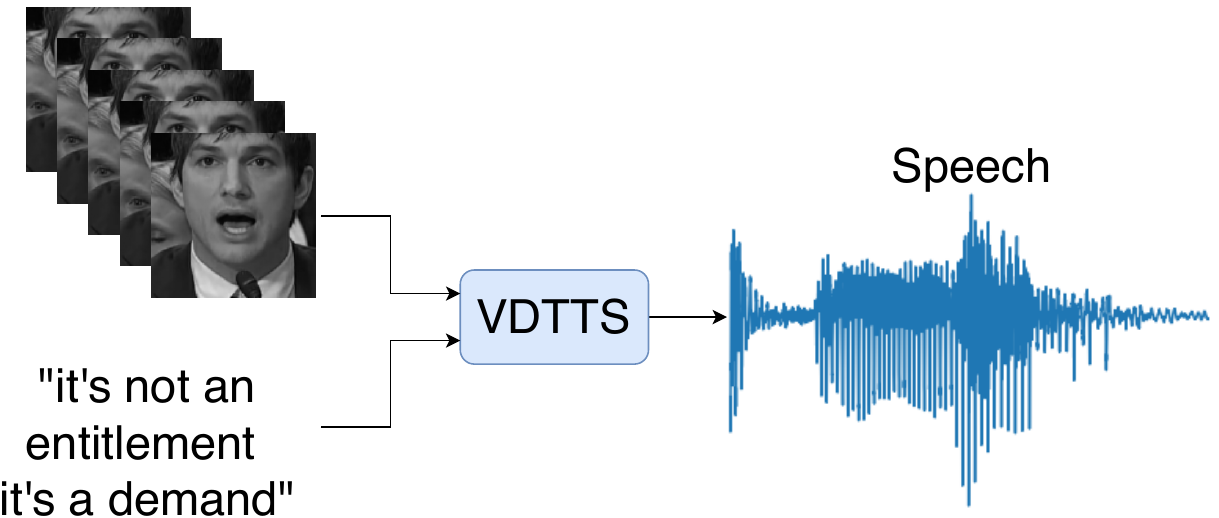}
\centering
\caption{Given a text and video frames of a speaker, VDTTS generates speech with prosody that matches the video signal.}
\label{fig:teaser}
\end{figure}

In this work, we present VDTTS, a visually-driven TTS model. Given text and corresponding video frames of a speaker speaking, our model is trained to generate the corresponding speech (see \cref{fig:teaser}).
As opposed to standard visual speech recognition models, which focus on the mouth region \cite{shillingford2018large}, we provide the full face to avoid potentially excluding information pertinent to the speaker's delivery.
This gives the model enough information to generate speech which not only matches the video but also recovers aspects of prosody, such as timing and emotion.
Despite not being explicitly trained to generate speech that is synchronized to the input video, the learned model still does so. 

Our model is comprised of four main components. Text and video encoders process the inputs, followed by a multi-source attention mechanism that connects these to a decoder that produces mel-spectrograms. A vocoder then produces waveforms from the mel-spectrograms.

We evaluate the performance of our method on GRID~\cite{cooke2006audio} as well as on challenging in-the-wild videos from VoxCeleb2~\cite{chung2018voxceleb2}. To validate our design choices and training process, we also present an ablation study of key components of our method, model architecture, and training procedure.

Demo videos are available on the project page,\footref{project_page} demonstrating video-speech synchronization, robustness to speaker ID swapping, and prosody. We encourage readers to take a look.

\vspace{1mm}
\noindent\textbf{Our main contributions are that we:} 
\begin{itemize} %[noitemsep,nolistsep]
	\item present and evaluate a novel visual TTS model, trained on a wide variety of open-domain YouTube videos;
	\item show it achieves state-of-the-art video-speech synchronization on GRID and VoxCeleb2 when presented with arbitrary unseen speakers; and
    \item demonstrate that our method recovers aspects of prosody such as pauses and pitch while producing natural, human-like speech.
\end{itemize}

\begin{figure*}[t]
\includegraphics[width=15cm]{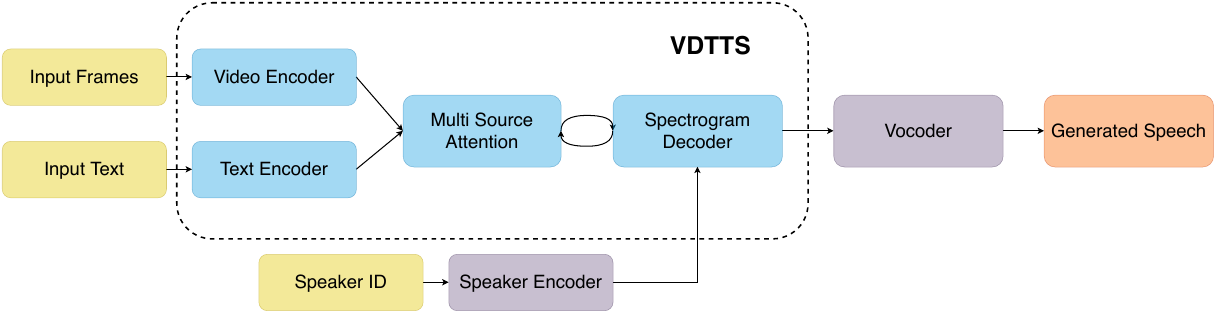}
\centering
\caption{The overall architecture of our model.
Colors: inputs: yellow, trainable: blue, frozen: purple, output: orange.
}
\label{fig:full_arch}
\end{figure*}
\section{Related work}
\label{sec:background}
\paragraph{Text-to-speech (TTS)}
engines which generate natural sounding speech from text, have seen dazzling progress in recent years. Methods have shifted from parametric models towards increasingly end-to-end neural networks~\cite{oord2016wavenet,wang2017tacotron}. This shift enabled TTS models to generate speech that sounds as natural as professional human speech~\cite{jia2021png}. Most approaches consist of three main components: an encoder that converts the input text into a sequence of hidden representations, a decoder that produces acoustic representations like mel-spectrograms from these, and finally a vocoder that constructs waveforms from the acoustic representations.
%The main difference between the different methods lies in how these three parts are designed and interact with their counterparts. 

Some methods including Tacotron and Tacotron 2 use an attention-based autoregressive approach~\cite{wang2017tacotron,shen2018natural,jia2018transfer}; followup work such as FastSpeech \cite{ren2019fastspeech, ren2020fastspeech2}, Non-Attentive Tacotron (NAT) \cite{shen2020non, jia2021png} and Parallel Tacotron \cite{elias2020parallel, elias2021parallel}, often replace recurrent neural networks with transformers.

Extensive research has been conducted on how to invert mel-spectrograms back into waveforms; since the former is a compressed audio representation, it is not generally invertible. For example, the seminal work of \citet{griffin1984signal} proposes a simple least-squares approach, while modern approaches train models to learn task-specific mappings that can capture more of the audio signal, including the approaches of WaveNet as applied to Tacotron 2~\cite{shen2018natural}, MelGAN~\cite{oord2016wavenet,kumar2019melgan}, or more recent work like WaveGlow \cite{prenger2019waveglow} which trains a flow-based conditional generative model, DiffWave \cite{kong2020diffwave} which propose a probabilistic model for conditional and unconditional waveform generation, or WaveGrad \cite{chen2020wavegrad} that make use of data density gradients to generate waveforms.
%removing the need for autoregression.
In our work, we use the fully-convolutional SoundStream vocoder~\cite{zeghidour2021soundstream}.

\paragraph{TTS prosody control}
%While the speech generated from a TTS engine might be as realistic as a human speaker, its style is often neutral or arbitrary. %, resulting in an unsatisfactory overall experience.%
\citet{skerry-ryan_towards_2018} define prosody as ``the variation in speech signals excluding phonetics, speaker identity, and channel effects."
Standard TTS approaches tend to be trained to produce neutral speech, due the difficulty of modeling prosody.

Great efforts have been made towards transferring or controlling the prosody of TTS audio.
\citet{wang_style_2018} created a style embedding by using a multi-headed attention module between the encoded input audio sequence and the global style tokens (GSTs). They trained a model jointly with the Tacotron model using the reconstruction loss of the mel-spectrograms.
%
%At inference time, the style embedding is created either from a certain text token to achieve style control, or from a different audio to achieve style transfer. 
At inference time, they construct the style embedding from the text to enable style control, or from other audio for style transfer. 

A Variational Auto-Encoder (VAE) latent representation of speaking style was used by \cite{zhang_learning_2019}. During inference time, they alter speaking style  by manipulating the latent embedding, or by obtained it from a reference audio. \citet{hsu_hierarchical_2018} used a VAE to create two levels of hierarchical latent variables, the first representing attribute groups, and the second representing more specific attribute configurations. This setup allows fine-grained control of the generated audio prosody including accent, speaking rate, etc.
%\todo{add FVAE-based approaches} 

\paragraph{Silent-video-to-speech} 
In this setup, a silent video is presented to a model that tries to generate speech consistent with the mouth movements, without providing text.
%no text nor transcripts are used.
%\todo{jiaye: does this include descriptive speech generation? if no, what's the difference to lip reading? answer by talremez@: Lip reading aims at generating text like an ASR model while here speech is generated likt in TTS. Its a semantic difference I think.}
Vid2Speech \cite{ephrat2017vid2speech} uses a convolutional neural network (CNN) that generates an acoustic feature for each frame of a silent video.
Lipper \cite{kumar2019lipper} use a closeup video of lips and produces text and speech,
while \cite{mira2021end} directly generated the speech without a vocoder.
\citet{prajwal2020learning} propose a speaker specific lip-reading model. 

\begin{table}[b]
    \footnotesize
    \centering
    \begingroup
    \setlength{\tabcolsep}{4pt}
    \begin{tabular}{l|rrrrll}
        \hline
                                                                 & Utts.  & Hrs.  & Vocab.    & Speakers & ID       & Source        \\ \hline\hline
        GRID \cite{cooke2006audio}                               & 34K    & 43    & 51        & 34       & \cmark   & Studio        \\
        LRS2 \cite{chung2017lipreadingsentences}                 & 47K    & 29    & 18K       & -        & \xmark   & BBC           \\
        LRS3 \cite{afouras2018lrs3}                              & 32K    & 30    & 17K       & 3.8K     & \xmark   & TED/TEDx      \\
        VoxCeleb2 \cite{chung2018voxceleb2}                      & 1M     & 2442  & \~{}35K*  & 6.1K     & \cmark   & YouTube       \\
        LSVSR \cite{shillingford2018large,yang_large-scale_2020} & 3M     & 3130  & 127K      & \~{}464K & \xmark   & YouTube       \\ \hline
    \end{tabular}
    \endgroup
    \caption{Audio-visual speech dataset size comparison in terms of number of utterances, hours, and vocabulary. Numbers are shown before processing;
    the resulting number of utterances we use for VoxCeleb2 and LSVSR are smaller.
    In \citet{yang_large-scale_2020}, LSVSR is called MLVD.
    (*VoxCeleb2 lacks transcripts, so we use an English-only automated transcription model \cite{park2020improved} to produce transcripts for training purposes, also used for vocabulary size measurement in this table.)
    }
    \label{tab:dataset}
\end{table}

\paragraph{Datasets}
For our task, we require triplets consisting of: a facial video, the corresponding speech audio, and a text transcript. The video and text are used as model inputs, whereas the speech audio is used as ground-truth for metrics and loss computation.
% For our task, we need triplets of facial video, corresponding speech audio, and text transcripts.

GRID is a standard dataset filmed under consistent conditions \cite{cooke2006audio}.
LRW \cite{chung2016lip} and LRS2 \cite{chung2017lipreadingsentences} are based on high-quality BBC television content, and LRS3 \cite{afouras2018lrs3} is based on TED talks; however, these datasets are restricted to academic use only.
VoxCeleb2 \cite{chung2018voxceleb2} and LSVSR \cite{shillingford2018large,yang_large-scale_2020},
being based on open-domain YouTube data, contain the widest range of people, types of content, and words. A comparison of dataset size appears in \cref{tab:dataset}.

In this work, we adopt GRID as a standard benchmark, and VoxCeleb2 and LSVSR due to their greater difficulty.

\paragraph{Automated dubbing}
A common approach to automated dubbing is to generate or modify the video frames to match a given clip of audio speech~\cite{yang_large-scale_2020,lahiri2021lipsync3d,suwajanakorn2017synthesizing,kumar2017obamanet,kr2019towards,song2020everybody,fried2019text,kim2019neural,jha2019cross}. This wide and active area of research uses approaches that vary from conditional video generation, to retrieval, to 3D models.
Unlike this line of work, we start from a fixed video and generate audio instead.

% This setup allows the model to focus on the alignment between the audio and the video frames rather than solely decode speech from lip motion . A main difference between our method and the ones listed above \todo{add some description on these works} is the ability to train it on large in-the-wild video corpus' such as VoxCeleb2. We also show we can use such a model on closed domain datasets such as GRID in addition to similar in-the-wild videos.
Recent visual TTS work uses both text and video frames to train a TTS model, much like our approach.
Concurrent work to ours \cite{lu2021visualtts, hu2021neural} take this approach, the former using GRID and the latter using just LRS2. Unlike our work, these approaches explicitly constrain output signal length and attention weights to encourage synchronization.

\section{Method}
\label{sec:method}
In this section, we describe the architecture of the proposed model and depict its components. Full architectural and training details are given in \cref{sec:arch_details} and \cref{sec:hyperparams} respectively.

\paragraph{Overview}
\cref{fig:full_arch} illustrates the overall architecture of the VTTS model. As shown, and similarly to \cite{ding2020textual}, the architecture consists of
(1) a video encoder, (2) a text encoder, (3) a speaker encoder, (4) an autoregressive decoder with a multi-source attention mechanism, and (5) a vocoder.
The method follows \cite{ding2020textual} using the combined \(L_1 + L_2 \) loss.
% , and the Teacher Forcing procedure.

Let $T_x$ and $T_y$ be the length of input video frame and phoneme sequences respectively. Let $D_w, D_h$ and $D_c$ be the width, height and the number of channels of the frames, $D_e$ the dimension of the phoneme embeddings, and $\mathcal P$ the set of phonemes.

We begin with an input pair composed of a source video frame sequence $x\in \mathbb{R}^{T_x \times D_w \times D_h \times D_c}$ and a sequence of phonemes $y \in \mathcal P^{T_y}$.
%, where \mbox{$x\in \mathbb{R}^{T_x \times D_w \times D_h \times D_c}$} and \mbox{$y\in \mathbb{R}^{T_y \times D_e}$} respectively. 
%Then $\bar y$ is embedded to produce a sequence of phoneme embeddings $y\in \mathbb{R}^{T_y \times D_e}$.
%

The video encoder receives a frame sequence as input, produces a hidden representation for each frame, and then concatenates these representations, i.e.,
\begin{equation}
    h_x = \textrm{VideoEncoder}(x) \in \mathbb{R}^{T_x \times D_m},
\label{eq:video_encoder}
\end{equation}
where \(D_m\) is the hidden dimension of the model.

Similarly, the text encoder receives the source phonemes and produces a hidden representation,
\begin{equation}
    h_y = \textrm{TextEncoder}(y)  \in \mathbb{R}^{T_y \times D_m}.
\label{eq:text_encoder}
\end{equation}
% \[  h_y = \textrm{TextEncoder}(y)  \in \mathbb{R}^{T_y \times D_m}.  \]

The speaker encoder maps a speaker 
to a 256-dimensional speaker embedding,
\begin{equation}
    d_i = \textrm{SpeakerEncoder}(\textrm{speaker}_i)  \in \mathbb{R}^{256}.
\label{eq:speaker_encoder}
\end{equation}

The autoregressive decoder receives as input the two hidden representations $h_x$ and $h_y$, and the speaker embedding $d_i$, and predicts the mel-spectrogram of the synthesized speech using the attention context,
\begin{equation}
    \hat{z}^t = \textrm{Decoder}(\hat{z}^{t-1}, h_x, h_y, d_i).
\label{eq:decoder}
\end{equation}

Finally, the predicted mel-spectrogram $[\hat{z}^1, \hat{z}^2, \ldots, \hat{z}^{T_z}]$ is transformed to a waveform using a frozen pretrained neural vocoder~\cite{zeghidour2021soundstream}.

\paragraph{Video encoder}
\label{chap:Video_Encoder}
Our video encoder is inspired by VGG3D as in~\cite{shillingford2018large}.
%More details about the specific architecture of the video encoder are given in \cref{sec:arch_details}.
However, unlike their work and similar lipreading work, we use a full face crop instead of a mouth-only crop to avoid potentially excluding information that could be pertinent to prosody, such as facial expressions.

\paragraph{Text encoder}
\label{chap:Text_Encoder}
Our text encoder is derived from Tacotron 2's \cite{shen2018natural} text encoder. Each phoneme is first embedded in a \(D_e\)-dimensional embedding space. Then the sequence of phoneme embeddings is passed through  convolution layers and a Bi-LSTM layer.
%For more specific details see \cref{sec:arch_details}.

\paragraph{Speaker encoder}
\label{chap:Speaker_Encoder}
In order to enable our model to handle a multi-speaker environment, we use a frozen, pretrained speaker embedding model~\cite{wan2018generalized} following \cite{jia2018transfer}.
When the speaker ID is provided in the dataset, as for GRID and VoxCeleb2, we generate embeddings per utterance and average over all utterances associated with the speaker, normalizing the result to unit norm.
For LSVSR the speaker identity is unavailable, so we compute the embedding per-utterance.
At test time, while we could use an arbitrary speaker embedding 
to make the voice match the speaker for comparison purposes, we use the average speaker embedding over the audio clips from this speaker.
We encourage the reader to refer to the project page \footref{project_page}, in which example videos demonstrate how VDTTS preforms when speaker voice embeddings are swapped between different speakers.

\paragraph{Decoder}
Our RNN-based autoregressive decoder is similar to the one proposed by \cite{shen2018natural}, and consists of four parts:
a \textit{pre-net}, a fully connected network reprojecting the previous decoder output onto a lower dimension before it is used as input for future time steps;
an attention module, in our case \textit{multi-source attention}, discussed later;
an LSTM core;
and a \textit{post-net} which predicts the final mel-spectrogram output.

The decoder receives as input the output sequences of the: video encoder \(h_x\), the text phoneme encoder \(h_y\) as well as the speaker embedding produced by the speaker encoder \(d_i\), and generates a mel-spectrogram of the speech signal \(\hat{z}^t\).
In contrast to \cite{shen2018natural}, which do not support speaker voice embeddings, we concatenate them to the output of the \textit{pre-net} to enable our model to be used in a multi-speaker environment, i.e: the input for the \textit{multi-source attention} at timestep \(t\) is
\begin{equation}
  q^t = \textrm{concat}(\textrm{PreNet}(\hat{z}^{t-1}), \space d_i).
\label{eq:decoder_msa_input}
\end{equation}
% Our multi-source attention differs from the attention of \cite{shen2018natural}, in the way we handle the speaker embedding \(d_i\), which we concatenate to the output of the \textit{pre-net}. 

\paragraph{Multi-source attention}
\label{chap:Multi_Source}
\begin{figure}[t]
\includegraphics[width=8cm]{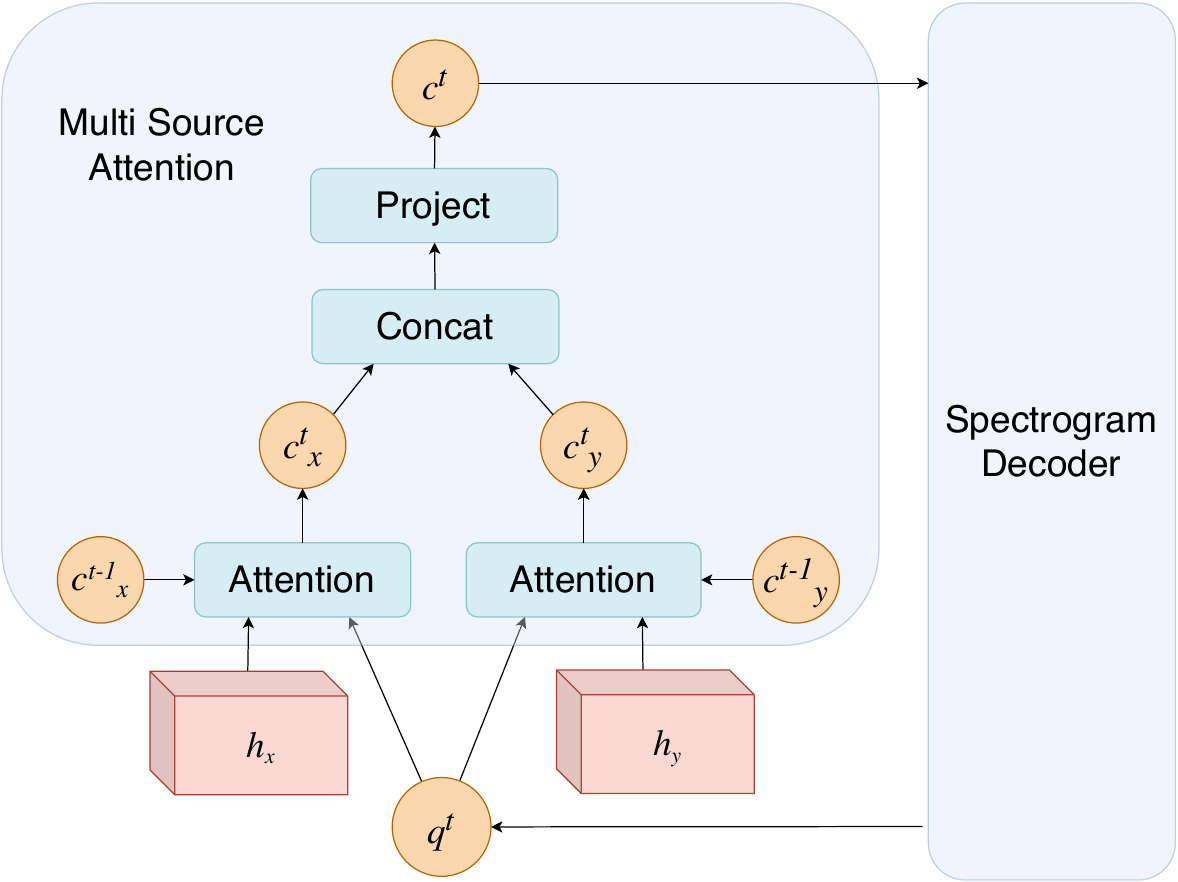}
\centering
\caption{The Multi Source Attention Mechanism.}
\label{fig:msa_diagram}
\end{figure}

A multi-source attention mechanism, similar to that of Textual Echo Cancellation \cite{ding2020textual}, allows selecting which of the outputs of the encoder are passed to the decoder in each timestep.

The multi-source attention, as presented in \cref{fig:msa_diagram}, has an individual attention mechanism for each of the encoders, without weights sharing between them.
At each timestep $t$, each attention module outputs an attention context,
% \begin{align*}
%   c_{x}^t &= \textrm{Atten}_x(q^t, c_{x}^{t-1}, h_x) \mbox{ and}\\
%   c_{y}^t &= \textrm{Atten}_y(q^t, c_{y}^{t-1}, h_y),
% \end{align*}
\begin{equation}
  c_{x}^t = \textrm{Att}_x(q^t, c_{x}^{t-1}, h_x); \,
  c_{y}^t = \textrm{Att}_y(q^t, c_{y}^{t-1}, h_y),
\label{eq:context_msa}
\end{equation}

where \(q^t\) is the output of the \textit{pre-net} layer of the decoder at timestep \(t\).

The input of the decoder at timestep $t$ is the projection of the concatenation of the two contexts described above via a linear layer,
\begin{equation}
  c^t = \textrm{Linear}([c_{x}^t, c_{y}^t]).
\label{eq:project_msa}
\end{equation}
While \cite{ding2020textual} aggregated the context vectors using summation, we found that a concatenation and projection work better in our setting as shown in \cref{chap:exp_ablation}.

%An important detail worth noting is that
We use a Gaussian mixture attention mechanism \cite{graves2013generating} for both modalities (video and text),
since it is a soft monotonic attention which is known to achieve better results for speech synthesis \cite{he2019robust, skerry2018towards,polyak2019attention}.

Full architectural details appear in \cref{sec:arch_details}.
% For the full details of the architecture above, see \cref{sec:arch_details}.

\section{Experiments}
\label{sec:experiments}

To evaluate the performance of the proposed video enhanced TTS model we conducted experiments on two very different public datasets: GRID \cite{cooke2006audio} and VoxCeleb2 \cite{chung2018voxceleb2}. 
GRID presents a controlled environment allowing us to test our method on high quality, studio captured videos with a small vocabulary, in which the same speakers appear in both the train and test sets. 
VoxCeleb2, however, is much more in-the-wild, therefore it is more diverse in terms of appearance (illumination, image quality, audio noise, face angles, etc.), and the set of speakers in the test set do not appear in the training set. This allows us to test the ability of the model to generalize to unseen speakers.

\subsection{Evaluation Metrics}
\label{sec:metrics}

We objectively evaluate prosodic accuracy, video-speech synchronization, and word error rate (WER).
We further evaluate synchronization subjectively with human ratings as described below.

Pitch (fundamental frequency, $F0$)
%\footnote{``Pitch'' commonly refers to $F0$, and $F0$ estimation methods are often referred to as ``pitch detection algorithms``.}
and voicing contours are computed using the output of the YIN pitch tracking algorithm \cite{decheveigne2002yinalgo} with a $12.5$ms frame shift. For cases in which the predicted signal is too short we pad using a domain-appropriate padding up to the length of the reference. If it is too long we clip it shorter.

In the remainder of this section we define and provide intuition for the metrics in the experimental section.

\subsubsection{Mel Cepstral Distortion (MCD$_K$) \cite{Kubichek93mcd}} is a mel-spectrogram distance measure defined as:
\begin{equation}
    \textrm{MCD} = \frac{1}{T} \sum_{t=0}^{T-1} \sqrt{ \sum_{k=1}^{K} (f_{t,k} - \hat{f}_{t,k})^2 },
\label{eq:mcd}
\end{equation}
where $\hat{f}_{t,k}$ and $f_{t,k}$ are the $k$-th Mel-Frequency Cepstral Coefficient (MFCC) \cite{tiwari2010mfcc} of the $t$-th frame from the reference and the predicted audio respectively. We sum the squared differences over the first $K=13$ MFCCs, skipping $f_{t,0}$ (overall energy). MFCCs are computed using a $25$ms window and $10$ms step size. 
%Lower MCD indicates that the predicted speech sounds more similar to the target speech.

\subsubsection{Pitch Metrics}
We compute the following commonly used prosody metrics over the pitch and voicing sequences produced from the synthesized and the ground-truth waveforms~\cite{skerry2018towards,sisman2020overview}.

\vspace{3mm}\noindent\textbf{F0 Frame Error (FFE) \cite{chu2009reducing}}
 measures the percentage of frames that either contain a $20\%$ pitch error or a voicing decision error.
\begin{equation} 
    {\textrm{FFE}} = \frac{\sum_{t}\mathds{1}[ | p_t - p'_t | > 0.2p_t]\mathds{1}[v_t=v'_t] + \mathds{1}[v_t \neq v'_t]}{T}
%\label{eq:pcc}   % commented because of many duplicate labels
\end{equation}
where $p$, $p'$ are the pitch, and $v$, $v'$ are the voicing contours computed over the predicted and ground-truth audio.

\vspace{3mm}\noindent\textbf{Gross Pitch Error (GPE) \cite{nakatani2008gpe}} measures the percentage of frames where pitch differed by more than $20\%$ on frames and voice was present on both the predicted and reference audio.
\begin{equation} 
    {\textrm{GPE}} = \frac{
      \sum_{t}\mathds{1}[ | p_t - p'_t |  > 0.2p_t]\mathds{1}[v_t = v'_t]
    }{
      \sum_{t}\mathds{1}[v_t = v'_t]
    }
%\label{eq:pcc}   % commented because of many duplicate labels
\end{equation}
where $p$, $p'$ are the pitch, and $v$, $v'$ are the voicing contours computed over the predicted and ground-truth audio.

\vspace{3mm}\noindent\textbf{Voice Decision Error (VDE) \cite{nakatani2008gpe}} measures the proportion of frames where the predicted audio is voiced differently than the ground-truth.
\begin{equation} 
    {\textrm{VDE}} = \frac{\sum_{t}\mathds{1}[v_t \neq v'_t]}{T}
%\label{eq:pcc}   % commented because of many duplicate labels
\end{equation}
where $v$, $v'$ are the voicing contours computed over the predicted and ground-truth audio.

% \vspace{3mm}\noindent\textbf{Pearson Correlation Coefficient (PCC)}
% \begin{equation} 
%     {\rho(p, p')} = \frac{\mbox{Cov}(p, p')}{\sigma_{p}\sigma_{p'}}
% %\label{eq:pcc}   % commented because of many duplicate labels
% \end{equation}
% where $\sigma_{p}$ and $\sigma_{p'}$ are the standard deviations of the predicted and ground-truth pitch sequences. We note that a higher PCC value represents better prosody retention performance. We use PCC to measure to measure pitch covariance which is less sensitive to overall voice pitch errors, i.e. where the entire predicted sample captures the pitch changes well but has an overall tone difference. 

\subsubsection{Lip Sync Error}
We use \emph{Lip Sync Error - Confidence} (LSE-C) and \emph{Lip Sync Error - Distance} (LSE-D) \cite{prajwal2020syncnet} to measure video-speech synchronization between the predicted audio and the video signal. The measurements are taken using a pretrained SyncNet model \cite{chung16syncnet}.
% while changing the minimal video duration to 0 and the minimal face size to 10 pixels. 

\subsubsection{Word Error Rate (WER)}
\label{chap:wer}

A TTS model is expected to produce an intelligible speech signal consistent with the input text. To measure this objectively, we measure WER as determined by an automatic speech recognition (ASR) model. To this end we use a state-of-the-art ASR model as proposed in \cite{park2020improved}, trained on the LibriSpeech \cite{panayotov2015librispeech} training set. The recognizer was not altered or fine-tuned. 

Since LSVSR is open-ended content, and out-of-domain compared to the audiobooks in LibriSpeech, the ASR performance may result in a high WER even on ground-truth audio. Thus, we only use the WER metric for relative comparison.
In \cref{sec:wer_discussion}, we compute the WER on predictions from a text-only TTS model trained on several datasets to establish a range of reasonable WERs; we confirm that a rather high WER is to be expected.

\subsubsection{Video-speech sync Mean Opinion Score (MOS)}
We measured video-speech synchronization quality with a 3-point Likert scale with a granularity of 0.5. Each rater is required to watch a video at least twice before rating it and a rater cannot rate more than 18 videos; each video is rated by 3 raters. Each evaluation was conducted independently; different models were not compared pairwise. The (averaged) MOS ratings are shown as a $90\%$ confidence interval.

In \cref{chap:exp_voxceleb} we rate a total of 200 videos each containing a unique speaker, while in \cref{chap:exp_grid} we chose 5 clips per speaker resulting in a total of 165 videos.
\begin{table*}
\centering
\small{
\begin{tabular}{l|ccccccccc}
\hline
                                    & MOS \ua               & LSE-C \ua   & LSE-D \da & WER \da    & MCD \da   & FFE \da   & GPE \da   & VDE \da \\ \hline \hline
\textsc{ground-truth} \cite{lu2021visualtts} &    -                  &     7.68    &   6.87    &     -      &      -    &     -     &      -    &    -    \\
\textsc{VisualTTS} \cite{lu2021visualtts}    &    -                  &     5.81    &   8.50    &     -      &      -    &     -     &      -    &    -    \\ \hline \hline

\textsc{ground-truth}                        &  2.68 $\pm$ 0.04      &     7.24    &   6.73    &     26\%   &      -    &     -     &      -    &    -    \\
\textsc{TTS-TextOnly} \cite{jia2021translatotron}&  1.51 $\pm$ 0.05      &     3.39    &   10.44   & \bb{19\%}  &   15.76   &   0.48    &   0.30    &   0.42  \\
\textsc{VDTTS-LSVSR}                          &  2.10 $\pm$ 0.06      &     5.85    &   7.93    &   55\%     &   12.81   &   0.37    &   0.21    &   0.32  \\
\textsc{VDTTS-GRID}                           &  \bb{2.55 $\pm$ 0.05} &  \bb{6.97}  & \bb{6.85} &   26\%     & \bb{7.89} & \bb{0.14} & \bb{0.07} &\bb{0.11}\\ \hline
\end{tabular}
}
\caption{\textbf{GRID evaluation.}
This table shows our experiments on the GRID dataset.
The top two rows present the numbers as they appear in \textsc{VisualTTS} \cite{lu2021visualtts}.
\textsc{ground-truth} shows the metrics as evaluated on the original speech/video.
\textsc{TTS-TextOnly} shows the performance of a vanilla text-only TTS model, while \textsc{VDTTS-LSVSR} and \textsc{VDTTS-GRID} are our model when trained on LSVSR and GRID respectively. While \textsc{VDTTS-GRID} archives the best overall performance, it is evident \textsc{VDTTS-LSVSR} generalizes well enough to the GRID dataset to outperform \textsc{VisualTTS} \cite{lu2021visualtts}.
See \cref{sec:metrics} for an explanation of metrics; arrows indicate if higher or lower is better. 
\label{tab:grid}}
\end{table*}
\begin{table*}[h]
\centering
\small{
\begin{tabular}{l|cccccccc}
\hline
                                                & MOS \ua               & LSE-C \ua & LSE-D \da & WER \da  & MCD \da    & FFE \da   & GPE \da   & VDE \da   \\ \hline \hline
\textsc{ground-truth}                           &  2.79 $\pm$ 0.03      &     7.00  &   7.51    &       -  &      -     &     -     &      -    &       -   \\
\textsc{TTS-TextOnly} \cite{jia2021translatotron}   &  1.77 $\pm$ 0.05      &     1.82  &  12.44    & \bb{4\%} &   14.67    & 0.59      &   0.38    & 0.42  \\
\textsc{VDTTS-VoxCeleb2}                        & \bb{2.50 $\pm$ 0.04}  & \bb{5.99} & \bb{8.22} &     48\% &   \bb{12.17}    & \bb{0.46} & 0.31 & \bb{0.30}  \\
\textsc{VDTTS-LSVSR}                            & 2.45 $\pm$ 0.04       & 5.92      &   8.25    &     25\% & 12.23 & \bb{0.46}      &   \bb{0.29}    & 0.31 \\ \hline

\end{tabular}
}

\caption{\textbf{VoxCeleb2 evaluation}.
%This table shows our performance on the VoxCeleb2 dataset.
\textsc{ground-truth} shows the synchronization quality of the original VoxCeleb2 speech and video.
\textsc{TTS-TextOnly} represents a vanilla text-only TTS model, while \textsc{VDTTS-VoxCeleb2} and \textsc{VDTTS-LSVSR} are our model when trained on  VoxCeleb2 and LSVSR respectively. By looking at the WER, it is evident \textsc{VDTTS-VoxCeleb2} generates unintelligible results, while \textsc{VDTTS-LSVSR} generalizes well to VoxCeleb2 data and produces better quality overall.
See \cref{sec:metrics} for an explanation of metrics; arrows indicate if higher or lower is better. 
\label{tab:voxceleb}}
\end{table*}
\subsection{Data preprocessing}
\label{sec:preprocessing}

Several preprocessing steps were conducted before training and evaluating our models, including audio filtering, face cropping and limiting example length.

We follow a similar methodology first proposed by \cite{shillingford2018large} while creating the LSVSR dataset. We limit the duration of all examples to be in the range of 1 to 6 seconds, and transcripts are filtered through a language classifier \cite{salcianu2018compact} to include only English. We also remove utterances which have less than one word per second on average, since they do not contain enough spoken content. We filter blurry clips and use a neural network \cite{chung2017lip} to verify that the audio and video channels are aligned. Then, we apply a landmarker as in \cite{schroff2015facenet} and keep segments where the face yaw and pitch remain within $\pm15^{\circ}$ and remove clips where an eye-to-eye width of less than 80 pixels. Using the extracted and smoothed landmarks, we discard minor lip movements and nonspeaking faces using a threshold filter. The landmarks are used to compute and apply an affine transformation (without skew) to obtain canonicalized faces. Audio is filtered \cite{kavalerov2019universal} to reduce non-speech noise.

We use this methodology to collect a similar dataset to LSVSR \cite{shillingford2018large}, which we use as our in-the-wild training set with $527,746$ examples, and also to preprocess our version of VoxCeleb2, only changing the maximal face angle to $30^{\circ}$ to increase dataset size.
Running the same processing as described above on VoxCeleb2 results in $71,772$ train, and $2,824$ test examples.
As for GRID which we use as our controlled environment, we do not filter the data, and only use the face cropping part of the aforementioned pipeline to generate model inputs.

\subsection{Controlled environment evaluation}
\label{chap:exp_grid}

In order to evaluate our method in a controlled environment we use the GRID dataset \cite{cooke2006audio}.
GRID is composed of studio video recordings of 33 speakers (originally 34, one is corrupt).
There are 1000 videos of each speaker, and in each video
a sentence is spoken with a predetermined ``GRID'' format.
The vocabulary of the dataset is relatively small and all videos were captured in a controlled studio environment over a green screen with little head pose variation.

We compare VDTTS to the recent VisualTTS \cite{lu2021visualtts} method using the same methodology reported by the authors. To that end, we take 100 random videos from each speaker as a test set. We use the remainder 900 examples per speaker as training data, and also for generating a lookup-table containing the speaker embedding, averaged and normalized per speaker, as explained in \cref{chap:Speaker_Encoder}.
At test time we present our models with video frames alongside the transcript and the average speaker embedding.

We evaluate our method using the metrics mentioned in \cref{sec:metrics}, and compare it to several baselines:
(1) \textsc{VisualTTS} \cite{lu2021visualtts};
(2) PnG NAT TTS zero-shot voice transferring model from \cite{jia2021translatotron}, a state-of-the-art TTS model trained on the LibriTTS \cite{zen2019libritts} dataset, denoted as \textsc{TTS-TextOnly};
(3) our model when trained over LSVSR (see \cref{sec:preprocessing}); and (4) our model trained on the GRID training set. 

Unfortunately, VisualTTS \cite{lu2021visualtts} did not provide their random train/test splits. Therefore, we report the original metrics as they appear in \cite{lu2021visualtts} alongside the numbers we found over our test set. Luckily, the two are comparable, as can be seen by the two rows in \cref{tab:grid} named \textsc{ground-truth}. 

The results appear in \cref{tab:grid}. Observe that, when trained on GRID, our method outperforms all other methods over in all metrics except WER.
Moreover, our model trained on LSVSR, as we will see in a later section, gets better video-speech synchronization results than VisualTTS, which was trained on GRID, showing that our ``in-the-wild'' model generalizes to new domains and unseen speakers.
%\bs{mention "later section", otherwise this won't read well sequentially}
% \bs{could this be a risky claim? GRID is much simpler than LSVSR}
\begin{table*}[ht]
\footnotesize
\setlength{\tabcolsep}{1pt}
\begin{tabular}{m{0.07\textwidth}cc}

&\begin{subfigure}{0.45\textwidth}\centering\includegraphics[width=1\columnwidth]{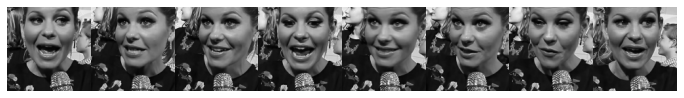}\end{subfigure}\vspace{0.015cm} &
\begin{subfigure}{0.45\textwidth}\centering\includegraphics[width=1\columnwidth]{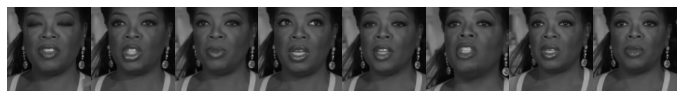}\end{subfigure}\vspace{0.015cm}\\
\textsc{GT} &\begin{subfigure}{0.35\textwidth}\centering\includegraphics[height=1.8cm,width=1\columnwidth]{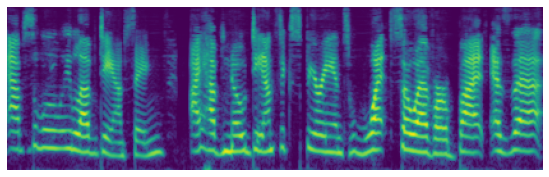}\end{subfigure}\vspace{0.015cm} &
\begin{subfigure}{0.35\textwidth}\centering\includegraphics[height=1.8cm,width=1\columnwidth]{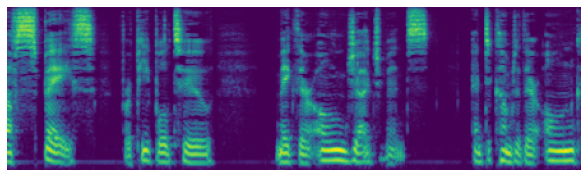}\end{subfigure}\vspace{0.015cm} \\
\textsc{VDTTS} &\begin{subfigure}{0.35\textwidth}\centering\includegraphics[height=1.8cm,width=1\columnwidth]{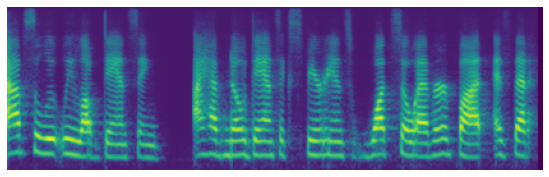}\end{subfigure}\vspace{0.015cm} &
\begin{subfigure}{0.35\textwidth}\centering\includegraphics[height=1.8cm,width=1\columnwidth]{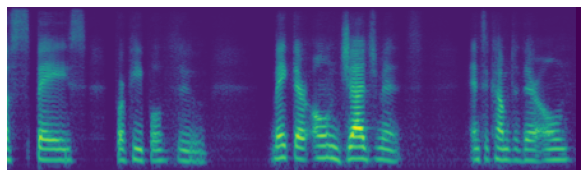}\end{subfigure}\vspace{0.015cm}\\
\textsc{TTS-TextOnly} \cite{jia2021translatotron} &
\begin{subfigure}{0.35\textwidth}\centering\includegraphics[height=1.8cm,width=1\columnwidth]{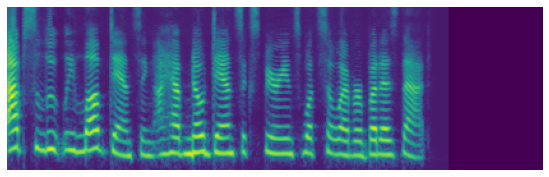}\end{subfigure}\vspace{0.015cm} &
\begin{subfigure}{0.35\textwidth}\centering\includegraphics[height=1.8cm,width=1\columnwidth]{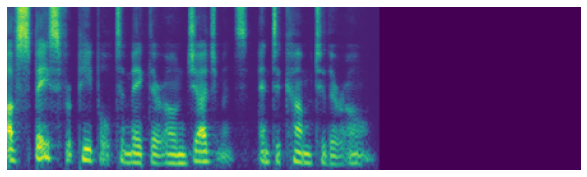}\end{subfigure}\vspace{0.015cm} \\
\textcolor{orange}{$F0$(\textsc{VDTTS})} \textcolor{cyan}{$F0$(\textsc{GT})} &
\begin{subfigure}{0.35\textwidth}\centering\includegraphics[height=1.8cm,width=1\columnwidth]{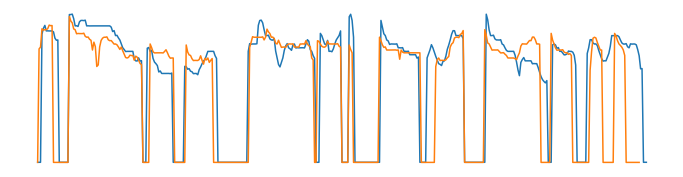}\end{subfigure}\vspace{0.015cm} &
\begin{subfigure}{0.35\textwidth}\centering\includegraphics[height=1.8cm,width=1\columnwidth]{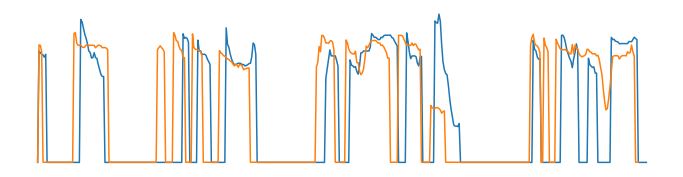}\end{subfigure}\vspace{0.015cm} \\
\textcolor{orange}{$F0$(\textsc{TTS-TextOnly})} \textcolor{cyan}{$F0$(\textsc{GT})} &
\begin{subfigure}{0.35\textwidth}\centering\includegraphics[height=1.8cm,width=1\columnwidth]{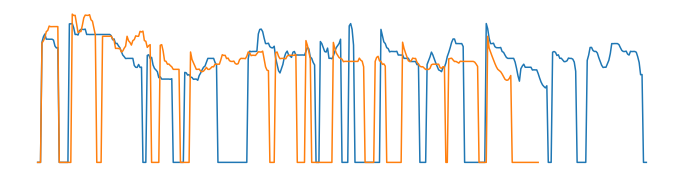}\end{subfigure}\vspace{0.015cm} &
\begin{subfigure}{0.35\textwidth}\centering\includegraphics[height=1.8cm,width=1\columnwidth]{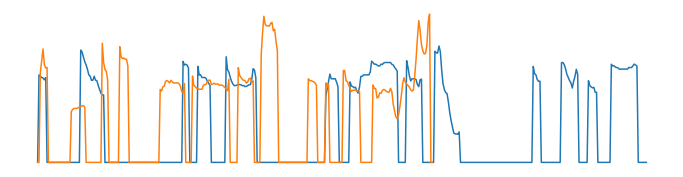}\end{subfigure}\vspace{-0cm} \\
& Time & Time \\
\vspace{0.2cm}
& (a) & (b) \\
\end{tabular}
\captionof{figure}{\textbf{Qualitative examples.} 
We present two examples (a) and (b) from the test set of VoxCeleb2 \cite{chung2018voxceleb2}. From top to bottom: input face images, ground-truth (\textsc{GT}) mel-spectrogram, mel-spectrogram output of \textsc{VDTTS}, mel-spectrogram output of a vanilla TTS model \textsc{TTS-TextOnly}, and two plots showing the normalized
%\footnote{Normalized pitch of a given waveform over time is computed by dividing the pitch by the mean pitch over frames where voice was present.}
%
pitch $F0$ (normalized by mean nonzero pitch, i.e. mean is only over voiced periods) of \textsc{VDTTS} and \textsc{TTS-TextOnly} compared to the ground-truth signal. For actual videos we refer the reader to the project webpage.
}
\label{fig:prosody}
\end{table*}
\subsection{In-the-wild evaluation}
\label{chap:exp_voxceleb}

In this section we evaluate VDTTS on the in-the-wild data from the test set of VoxCeleb2 \cite{chung2018voxceleb2}. This is an open-source dataset made of in-the-wild examples of people speaking and is taken from YouTube.
We preprocess the data as described in \cref{sec:preprocessing}. Since this data is not transcribed, we augment the data with transcripts automatically generated using \cite{park2020improved}, yielding $2{,}824$ high quality, automatically transcribed test videos. We create a speaker embedding lookup table by averaging and normalizing the speaker voice embeddings from all examples of the same speaker.

As a baseline we again use \textsc{TTS-TextOnly}, a text-only TTS model from \cite{jia2021translatotron}. 
The results are shown in \cref{tab:voxceleb}.

Initially we trained our model on the train set of VoxCeleb2, called \textsc{VDTTS-VoxCeleb2}.
%Unfortunately this model yielded a relatively poor WER score of $48\%$.
%As discussed in detail in \cref{chap:wer}, assuming a good ASR model a high WER suggests difficult-to-comprehend audio,
%and we suspect that noisy automated transcripts was a cause of poor training.
%Thus, we decided to train our model on a larger dataset with human generated transcripts.
Unfortunately, as can be seen by the high WER of $48\%$, the model produced difficult-to-comprehend audio. We hypothesized that noisy automated transcripts were the culprit, so trained the model on an alternative in-the-wild dataset with human generated transcripts, LSVSR, we denote this model by \textsc{VDTTS-LSVSR}. %
As hypothesized, this leads to a great improvement in WER and reduced the error to only $24\%$ while leaving most other metrics comparable. For more details refer to \cref{sec:wer_discussion}.

For qualitative examples of \textsc{VDTTS-LSVSR} we refer the reader to \cref{sec:exp_prosody}.
\subsection{Prosody using video}
\label{sec:exp_prosody}
We selected two inference examples from the test set of VoxCeleb2 to showcase the unique strength of VDTTS, which we present in \cref{fig:prosody}. In both examples, the video frames provide clues about the prosody and word timing. Such visual information is not available to the text-only TTS model, \textsc{TTS-TextOnly}~\cite{jia2021translatotron}, to which we compare.

In the first example (see \cref{fig:prosody}(a)), the speaker talks at a particular pace that results in periodic gaps in the ground-truth mel-spectrogram.
The VDTTS model preserves this characteristic and generates mel-spectrograms that are much closer to the ground-truth than the ones generated by \textsc{TTS-TextOnly} without access to the video.

Similarly, in the second example (see \cref{fig:prosody}(b)), the speaker takes long pauses between some of the words.
This can be observed by looking at the gaps in the ground-truth mel-spectrogram.
These pauses are captured by  VDTTS and are reflected in the predicted result below, whereas the mel-spectrogram of \textsc{TTS-TextOnly} does not capture this aspect of the speaker's rhythm.

We also plot $F0$ charts to compare the pitch generated by each model to the ground-truth pitch. In both examples, the $F0$ curve of VDTTS fits the ground-truth much better than the \textsc{TTS-TextOnly} curve, both in the alignment of speech and silence, and also in how the pitch changes over time.

To view the videos and other examples, we refer the reader to the project page\footref{project_page}. 

\begin{table}[tb]
\footnotesize
\centering
\begin{tabular}{@{\hskip 2pt}l@{\hskip 2pt}|@{\hskip 2pt}c@{\hskip 2pt}c@{\hskip 2pt}c@{\hskip 2pt}c@{\hskip 2pt}c@{\hskip 2pt}c@{\hskip 2pt}c@{\hskip 2pt}}
\hline
                        & \scriptsize{LSE-C} \ua     & \scriptsize{LSE-D} \da   & \scriptsize{WER} \da     & \scriptsize{MCD} \da   &    \scriptsize{FFE} \da    & \scriptsize{GPE} \da   & \scriptsize{VDE} \da   \\  \hline
Full VDTTS                   &     \bb{5.92} &   \bb{8.25} & \bb{25\%}   & 12.23 &   \bb{0.46}        &   \bb{0.29}    & \bb{0.31} \\ 
VDTTS-no-sp-emb    &     1.49      &   12.14     & 27\%        &   14.5    &   0.67        &   0.43    &  0.37     \\ 
VDTTS-small            &     1.48      &   12.45     & 38\%        &   14      &   0.6         &   0.4     &  0.43     \\ 
VDTTS-sum-att       &     5.74      &   8.47      & 28\%        &   \bb{12.22}   &   \bb{0.46}   & \bb{0.29} &  \bb{0.31}     \\ 
VDTTS-no-text           &     5.90      &   8.28      & 98\%        &   12.99   &   0.53        &   0.35    &  0.35      \\ 
VDTTS-no-video          &     1.44      &   12.62     & 27\%        &   14.36   &   0.58        &   0.34    &  0.47     \\ 
VDTTS-video-len          &     1.58      &   12.37     & 28\%        &   13.98   &   0.59        &   0.37    &  0.42     \\ 
VDTTS-mouth          &     5.51      &   8.59     & 29\%        &   12.24   &   0.52        &   0.41    &  \bb{0.31}     \\ 
\end{tabular}

\caption{\textbf{Ablation study}, showing different variations of the VDTTS model and hence the contribution of these components to the performance of VDTTS. See \cref{chap:exp_ablation} for a detailed explanation of the different models, and \cref{sec:metrics} for definitions of metrics. Arrows indicate if higher or lower is better.
\label{tab:ablation}}
\end{table}
\subsection{Ablation}
\label{chap:exp_ablation}
%\hassid{Let's agree on 1 naming convention for the models below: "x" (like in the table and the first pargraph) or "VDTTS-x" like in the rest of the paragraph.}

In this section we conduct an ablation study to better understand the contribution of our key design choices.

Results are presented in \cref{tab:ablation} using the following abbreviations for the models:
(1) \emph{VDTTS-no-sp-emb}: VDTTS without the use of a speaker embedding. Although unlikely, this version could possibly learn to compensate for the missing embedding using the person in the video. 
(2) \emph{VDTTS-small}: VDTTS with smaller encoders and decoder, with $D_m=512$ as in \cite{shen2018natural}.
(3) \emph{VDTTS-sum-att}: VDTTS using a summation (as in \cite{ding2020textual}) instead of concatenation in the Multi Source Attention mechanism.
(4) \emph{VDTTS-no-text}: VDTTS without text input, can be thought of as a silent-video-to-speech model.
(5) \emph{VDTTS-no-video}: VDTTS without video input, can be thought of as a TTS model.
(6) \emph{VDTTS-video-len}: VDTTS trained with empty frames, used as a baseline of a TTS model which is aware of the video length. 
(7) \emph{VDTTS-mouth}: VDTTS which is trained and evaluated on the mouth region only (as in most speech recognition models). 

\emph{VDTTS-no-sp-emb} performs poorly on the video-speech synchronization metrics LSE-C and LSE-D,
%we hypothesis this due to the limited capacity of the video encoder, as it is required to learn both voice properties to compensate for the lack of a speaker embedding, as well as capture mouth and face movements for audio-video synchronization.
likely due to underfitting since the model is unable to infer the voice of the speaker using only the video.

Looking at \emph{VDTTS-small}, makes it evident that increasing $D_m$ beyond what was originally suggested by \citet{ding2020textual} is required.

Another interesting model is \emph{VDTTS-no-text}, which has access only to the video frame input without any text. In terms of video-speech synchronization it is on par with the full model for LSE-C and LSE-D, but fails to produce words as can be seen by its high WER.
Intriguingly, outputs from this model look clearly synchronized, but sounds like English babbling, as can be seen in the examples on the project page\footref{project_page}.
On one hand, this shows that the text input is necessary in order to produce intelligible content,
and on the other hand it shows the video is sufficient for inferring synchronization and prosody without having access to the underlying text. 
%
% Furthermore, in both this model and the full one, the synchronization is learnt without any explicit loss or  constraint to encourage it, suggesting that the ``easiest" solution for the model to learn is to infer prosody visually, rather than modeling it from the text.

Although it seems that \emph{VDTTS-video-len} shows similar results to the \emph{VDTTS-no-video} model, the former produces speech signal which corresponds to the original scene length (as desired), which the latter does not.

Lastly, the \emph{VDTTS-mouth} performs a bit worse than the full model, which suggests that the use of the full face crop is indeed beneficial to the model.

\section{Discussion and Future Work}

In this paper we presented VDTTS, a novel visually-driven TTS model that takes advantage of video frames as an input and generates speech with prosody that matches the video signal.
Such a model can be used for post-sync or dubbing, producing speech synchronized to a sequence of video frames.
Our method also naturally extends to other applications such as low-quality speech enhancement in videos and audio restoration in captioned videos.

VDTTS produces near ground-truth quality on the GRID dataset.
On open-domain ``in-the-wild'' evaluations, it produces well-synchronized outputs approaching the video-speech synchronization quality of the ground-truth and performs favorably compared to alternate approaches.

% KEY HERE: put bad/shortcomings near good
Intriguingly, VDTTS is able to produce video-synchronized speech without any explicit losses or constraints to encourage this, suggesting complexities such as synchronization losses or explicit modeling are unnecessary.
Furthermore, we demonstrated that the text and speaker embedding supply the speech content and voice, while the prosody is produced by the video signal.
Our results also suggest that the ``easiest" solution for the model to learn is to infer prosody visually, rather than modeling it from the text. 

%In the context of synthesis it is important to address the potential for misuse by generating convincingly false audio. Since VDTTS is trained using video and text pairs in which the video corresponds exactly to the text being said, it cannot synthesize from arbitrary text, thus making it less prone to misuse and the generation of fake content. 
In the context of synthesis it is important to address the potential for misuse by generating convincingly false audio. Since VDTTS is trained using video and text pairs in which the speech pictured in the video corresponds to the text spoken, synthesis from arbitrary text is out-of-domain, making it unlikely to be misused.

% For future work, it remains to be seen how VDTTS chould be modified to generate video-synchronous audio for arbitrary text input text, which would be a powerful tool for performing, for instance, translation dubbing.

%-------------------------------------------------------------------------

%%%%%%%%% REFERENCES
\clearpage
{\small
\bibliographystyle{IEEEtranN}
\bibliography{egbib}
}

\clearpage
\appendix
% \section*{Appendix}
\section{Detailed model architecture}
\label{sec:arch_details}

% \begin{table*}[h]
% \centering
\begin{small}
\begin{tabular}{l||c|cc|}
\hline
                            
Video Encoder &  Input size  & \(D_w=D_h=128, D_c=1\)
\tabularnewline\cline{2-3} & Max Pooling per layer &            (F, T, T, T, T)
\tabularnewline\cline{2-3} & Number of output channels per layer &            (64, 128, 256, 512, 512)       
\tabularnewline\cline{2-3} & Stride per layer         &            (2, 1, 1, 1, 1)      
\tabularnewline\cline{2-3} & Kernel size (for all layers)         &            (3, 3, 3)     
\tabularnewline\cline{2-3} & Activation (for all layers)         &            ReLU      
\tabularnewline\cline{2-3} & Normalization (for all layers)         &            Group Norm \\  
\hline
\hline

Text Encoder & Input size  & \(D_e=512\)
\tabularnewline\cline{2-3} & Conv layers \(\times\) 3         &            2048 5\(\times\)1 kernel with 1\(\times\)1 stride      
\tabularnewline\cline{2-3} & Activation (for all Conv layers)         &            ReLU      
 \tabularnewline\cline{2-3} & Bi-LSTM  &            1024-dim per direction 
 \tabularnewline\cline{2-3} & Normalization (for all Conv layers)         &            Batch Norm    \\
\hline
\hline
Multi Source Attention & Attention input size & $D_m=2048$
\tabularnewline\cline{2-3} & GMM attention (per source) &  128-dim context
\tabularnewline\cline{2-3} & Linear Projection         &  Fully connected layer  \\
\hline
\hline
Decoder &PreNet &           2 fully connected layers with 256 neurons and ReLU act.
\tabularnewline\cline{2-3} & LSTM \(\times\) 2       &            1024-dim      
 \tabularnewline\cline{2-3} & Bi-LSTM  &            1024-dim per direction 
 \tabularnewline\cline{2-3} & PostNet        &            5 conv layers with 512 5\(\times\)1 kernel with 1\(\times\)1 stride 
 \tabularnewline &         &     and TanH act. 
 \tabularnewline\cline{2-3} & Normalization (for all Decoder layers)         &            Batch Norm  
 \tabularnewline\cline{2-3} & Teacher forcing prob         &            1.0  \\
\hline
\end{tabular}
\end{small}
% \label{tab:arch_details}
% \caption{Detailed specifications of VDTTS model architecture.}
% \end{table*}

\section{Training hyperparameters}
\label{sec:hyperparams}  % TODO: rename to section

% \begin{table*}[h]
% \centering
\begin{small}
\begin{tabular}{l||c|cc|}
\hline
                            
Training &  learning rate  &            0.0003
\tabularnewline\cline{2-3} & learning rate scheduler type &            Linear Rampup with Exponential Decay 
\tabularnewline\cline{2-3} & scheduler decay start &           40k steps
\tabularnewline\cline{2-3} & scheduler decay end &           300k steps
\tabularnewline\cline{2-3} & scheduler warm-up &           400 steps  
\tabularnewline\cline{2-3} & batch size &           512\\ 
\hline
\hline

Optimizer  & optimizer details &            Adam with \(\beta_1 = 0.9, \beta_2 = 0.999\) \\
\hline
\hline
Regularization  & L2 regularization factor &  1e-06 \\
\hline
% \hline
% Hardware &      TPU version  & TPUv3
% \tabularnewline\cline{2-3} & Number of TPUs       &            64     
%  \tabularnewline\cline{2-3} & Batch size per TPU core &            4 
%  \tabularnewline\cline{2-3} & Total batch size        &           512 \\
% \hline
\end{tabular}
\end{small}
% \label{tab:hyperparams}
% \caption{Hyperparameters for the training procedure.}
% \end{table*}

\clearpage

\section{Word error rate discussion}
\label{sec:wer_discussion}

As explained in \cref{chap:exp_voxceleb}, our VoxCeleb2 transcripts are automatically generated and thus contain transcription errors. As a result one can expect the WER for models trained on this data to be non-zero.
In order to validate this hypothesis, that the result of such noisy data leads to a non-zero WER, we trained a version of the our model that accepts only text as input (without silent video), denoted as \textsc{TTS-our}. \textsc{TTS-our} was trained twice, once on the LibriTTS \cite{zen2019libritts} dataset, and a second time when using our in-the-wild LSVSR dataset. 
When looking at \cref{tab:wer} it is clear that when trained on LibriTTS this model achieves a low WER of 7\%, while the same model when trained on in-the-wild dataset get a WER of 27\%. This suggests that a WER in the region of $[20\%,30\%]$ should be expected when using LSVSR.

That being said, we believe reporting WER is valuable as a sanity check for noisy datasets, specially when trying to capture more than just the words.

\begin{table}[ht]
\centering
\begin{tabular}{l|c}
\hline
Training data   & WER  \\ \hline \hline
\textsc{LibriTTS}        & 7\%  \\
\textsc{LSVSR}           & 27\%  \\
\hline
\end{tabular}
\caption{Comparison of WER on the VoxCeleb2 test set for our text only TTS model (\textsc{TTS-our}) when trained on different datasets. 
\label{tab:wer}}
\end{table}

\end{document}